\documentclass[letterpaper, 10 pt, conference]{ieeeconf}  % Comment this line out if you need a4paper
\usepackage{amsmath}
\usepackage{subcaption}
\usepackage{graphicx}
\usepackage{multirow}
\usepackage{booktabs}
\usepackage{amsmath,amssymb,amsfonts}
\usepackage{xcolor}
\IEEEoverridecommandlockouts                              % This command is only needed if
\title{\LARGE \bf
From Sign Language Generation to Humanoid Execution: Vision–Language Guided Retargeting with Collision Mitigation
}

\author{Nabeela Khan$^{1,2}$, Bowen Wu$^{2,3}$, Runwu Shi$^{1}$, Benjamin Yen$^{4,1}$, Takeshi Ashizawa$^{1}$,\\ Carlos Toshinori Ishi$^{2,3}$, Takashi Minato$^{2,3}$ and Kazuhiro Nakadai$^{1}$% <-this % stops a space
\thanks{This paper has been accepted for presentation at the 2026 IEEE/RSJ International Conference on Intelligent Robots and Systems (IROS 2026).}%
\thanks{*This work was supported in part by JST Moonshot R\&D under Grant JPMJMS2011, JSPS KAKENHI Grant No. 26H02525, and ROIS NII Open Collaborative Research 2026-261S04-24183. }% <-this % stops a space
\thanks{$^{1}$Department of Systems and Control Engineering, Institute of Science Tokyo, Tokyo, Japan
        {\tt\small nabeela@ra.sc.e.titech.ac.jp}}%
\thanks{$^{2}$RIKEN Guardian Robot Project, Kyoto, Japan}%
\thanks{$^{3}$ATR Hiroshi Ishiguro Laboratories, Kyoto, Japan}%
\thanks{$^{4}$RIKEN Center for Biosystems Dynamics Research, Kobe, Japan}%
}

\begin{document}

\maketitle

\begingroup
\renewcommand{\thefootnote}{}
\footnotetext{© 2026 IEEE. Personal use of this material is permitted.
Permission from IEEE must be obtained for all other uses, including
reprinting/republishing this material for advertising or promotional purposes,
creating new collective works for resale or redistribution to servers or lists,
or reuse of any copyrighted component of this work in other works.}
\addtocounter{footnote}{-1}
\endgroup

\thispagestyle{empty}
\pagestyle{empty}

%%%%%%%%%%%%%%%%%%%%%%%%%%%%%%%%%%%%%%%%%%%%%%%%%%%%%%%%%%%%%%%%%%%%%%%%%%%%%%%%
\begin{abstract}

Recent sign language generation (SLG) systems increasingly output dense 3D body representations, which better preserve full-body kinematics and geometry for downstream embodiment on humanoid robots. However, these generated motions frequently exhibit self-intersections such as hand–hand and hand–torso penetration. While such artifacts may be tolerated in offline rendering, they become critical in humanoid execution as they lead to infeasible inverse-kinematics (IK) solutions, collisions, and unstable retargeted trajectories. We present a system-level framework that bridges SLG outputs to humanoid joint-space execution via two components. First, we introduce a volumetric SMPL-X collision-mitigation module that projects generated signing motions toward physically plausible configurations while minimally deviating from the original trajectory. Second, we propose a vision–language–guided retargeting algorithm built on an IK backbone: a VLM serves as a visual critic over rendered humanoid motion, identifies embodiment-specific failure modes, and triggers targeted task-space corrections. Our results highlight collision handling and perception-guided refinement as key missing components for reliable humanoid signing.
\end{abstract}

%%%%%%%%%%%%%%%%%%%%%%%%%%%%%%%%%%%%%%%%%%%%%%%%%%%%%%%%%%%%%%%%%%%%%%%%%%%%%%%%
\section{INTRODUCTION}

Sign language (SL) is a primary communication modality for Deaf and Hard-of-Hearing (DHH) communities\cite{yin2021including}, and enabling robots to express SL offers a direct path toward accessible human–robot communication in public spaces, service contexts, and assistive interaction \cite{qiao2025signbot, khan2025towards}. While sign language processing has advanced rapidly in recognition and translation, sign language generation (SLG)\cite{tan2024review}, the synthesis of sign motions from spoken or written language remains challenging, particularly when moving from offline animation to physical embodiment on a humanoid platform.

Recent SLG models increasingly aim to produce full-body 3D motion rather than sparse upper-body keypoints \cite{khan2025signflow}. Dense human-body representations such as SMPL-X \cite{pavlakos2019expressive} are attractive because they encode whole-body kinematics and a coherent body surface, enabling consistent joint coordination, more realistic articulation, and downstream embodiment pipelines that require explicit geometry. In contrast, sparse skeletons often discard surface information that is critical when transferring motion to a robot, where feasibility depends on collisions, joint limits, and realistic limb trajectories. This shift toward dense representations, therefore, brings SLG closer to physical execution, but it also exposes a gap that is easy to overlook in purely visual evaluation.

A key failure mode is self-collision, including hand–hand and hand–torso interpenetration, which appears frequently in generated sign motions. In offline rendering, mild penetration may be visually tolerable or may pass unnoticed under typical qualitative inspection. For humanoid execution, however, self-collision becomes a first-order constraint: penetrations can yield infeasible inverse-kinematics (IK) solutions, force discontinuous corrections, and induce collision events or unstable control after retargeting \cite{araujo2025retargeting}. Importantly, these artifacts are not merely a “retargeting problem.” They often originate upstream in SLG due to imperfect motion priors and learning objectives that do not explicitly enforce physical plausibility. As a result, even a strong retargeting backbone can inherit penetrations and amplify them into execution failures.

At the same time, retargeting SL motion to humanoids is underexplored compared to retargeting for locomotion, whole-body imitation, or generic tracking. Existing humanoid retargeting pipelines typically prioritize reproducing reference kinematics under robot constraints\cite{jiang2024harmon}, but SL introduces additional challenges: meaning depends on subtle hand trajectories, relative placement in signing space, and coordination between torso and arms. A retargeter must therefore balance constraint satisfaction (joint limits, collisions) with motion fidelity to preserve communicative intent. This makes SLG-to-humanoid retargeting a distinct robotics problem rather than a straightforward reuse of animation retargeting tools.

A natural alternative to retargeting generated motion would be a precomputed dictionary of collision-free, robot-specific sign trajectories that are stitched together at runtime. However, SLs are productive rather than a fixed inventory of isolated tokens: the realization of a sign varies with sentence context through spatial referencing, coarticulation with neighboring signs, and grammatical use of the signing space, so continuous signing cannot in general be reproduced by concatenating canonical dictionary entries without degrading naturalness and meaning. Moreover, a trajectory dictionary is tied to a specific embodiment and must be re-authored for every robot morphology, whereas SLG models already produce open-vocabulary, context-dependent motion in a shared human-body representation. A generative retargeting pipeline therefore scales across sentences and platforms; conversely, our framework can itself be used as an automated offline tool for constructing validated, robot-specific sign trajectory libraries where a fixed vocabulary suffices.

In this work, we address this gap with a framework that converts generated sign motions to humanoid joint trajectories through two complementary components: a collision-mitigation module based on volumetric SMPL-X self-intersection modeling~\cite{mihajlovic2025volumetricsmpl}, which projects the generated motion toward physically plausible configurations while constraining deviation from the original trajectory and maintaining temporal smoothness; and a vision--language model (VLM) guided retargeting algorithm built on an IK backbone, in which the VLM acts as a visual critic on rendered humanoid motion and triggers targeted task-space corrections through a small set of interpretable primitives.

Our primary contribution is system-level: rather than proposing a new motion-generation algorithm, we identify the failure modes that arise specifically at the point of physical embodiment and integrate collision mitigation and perception-guided refinement into a unified, practical SLG-to-humanoid pipeline. We summarize our contributions as follows:

\begin{itemize}
    \item We formulate SLG-to-humanoid transfer as an embodiment pipeline in which geometric artifacts inherited from generated SMPL-X motion must be resolved before robot retargeting, and instantiate this with a volumetric SMPL-X collision-mitigation module with trajectory-preserving and temporal smoothness regularization.
    \item We introduce an IK-based retargeting pipeline augmented with a VLM visual critic that identifies failure modes from rendered humanoid motion and triggers targeted corrective primitives.
\end{itemize}

\section{Related Work}
\subsection{Sign Language Generation}

Early SLG research largely relied on \textit{glosses}\footnote{A gloss is a label representing an isolated sign, often used as an intermediate symbolic representation.} as an intermediate representation, following a text$\rightarrow$gloss$\rightarrow$pose (or video) pipeline \cite{huang2021towards,saunders2022signing,tang2022gloss,xie2024g2p,walsh2024select,tang2025gloss,zuo2024simple,xie2024sign,zhang2025towards}. These systems were commonly assessed with keypoint- or pose-based metrics and qualitative visual inspection \cite{saunders2020progressive,yin2024t2s}.
More recent work has pushed toward end-to-end, gloss-free SLG and toward 3D body representations that better capture full-body coordination and signing-space geometry \cite{saunders2021continuous,khan2025multigau, khan2025end}. In particular, approaches that predict dense 3D human-body motion (e.g., SMPL/SMPL-X pose parameters or mesh-consistent kinematics) are increasingly attractive for downstream embodiment because they provide a coherent articulation model and an explicit body surface that can be rendered or transferred to other embodiments \cite{zuo2025signs,baltatzis2024neural}.
Despite this progress, most SLG work still evaluates outputs primarily in the avatar rendering domain, where geometric artifacts, such as mild self-interpenetration or near-contacts, may not be penalized by standard metrics (Sec.~I). Our work complements SLG by addressing this gap: we treat SLG as an upstream generator and focus on the missing \emph{embodiment layer} needed for reliable humanoid execution.

\subsection{Self-Collision}
\label{subsec:rw_collision}

Self-collision has long been recognized as a fundamental challenge in articulated human motion synthesis and character animation \cite{mihajlovic2025volumetricsmpl}. In graphics and vision, collision handling is commonly addressed using geometric penalties, signed distance fields (SDFs), capsule/primitive-based approximations, or volumetric/implicit representations that enable differentiable intersection tests and gradient-based correction \cite{pavlakos2019expressive,fieraru2021remips,mihajlovic2022coap,davydov2024cloaf,macklin2020local}. In learning-based motion generation, collision artifacts are especially common when objectives prioritize reconstruction or realism without explicit physical constraints, and when the model must resolve one-to-many motion realizations.

As noted in Sec.~I, such penetrations directly compromise downstream robot execution~\cite{koptev2021real}. Related observations also arise in humanoid retargeting, where imperfect tracking, scaling, or embodiment mismatch can introduce self-penetration and physically infeasible configurations \cite{araujo2025retargeting}. These issues are particularly acute for SL motions, which frequently involve rapid bimanual interactions close to the torso. Motivated by this gap between visually plausible motion and robot-executable motion, we introduce a lightweight collision-mitigation module for SLG outputs using a volumetric self-intersection formulation (SMPL-X-based) that projects motions toward physically plausible configurations while limiting deviation from the original signing trajectory.

\subsection{VLM-Guided Humanoid Retargeting}
\label{subsec:rw_vlm_retarget}

Humanoid signing has been explored as an assistive capability in human--robot interaction, but deploying sign motions on physical robots remains challenging due to embodiment constraints and the need for reliable motion transfer from human/avatar representations \cite{qiao2025signbot}. A key technical bottleneck is the gap between human/avatar motion and robot-executable motion: humanoid robots must satisfy kinematic limits, joint couplings, link geometry, collision constraints, and reachable workspace, and therefore require dedicated retargeting and control mechanisms. Recent work has begun integrating these components into end-to-end sign-language-centered pipelines. For instance, SignBot \cite{qiao2025signbot} includes motion retargeting to robot kinematics and reports simulation and real-robot experiments across multiple robots/datasets.
In parallel, the broader robotics community has started leveraging vision--language models (VLMs) as perceptual and commonsense critics for embodied behavior. Recent methods show that VLMs can evaluate rendered or camera observations, detect failure modes that are difficult to specify as analytic costs, and provide corrective guidance for planning or control \cite{zhao2024vlmpc,duan2024aha}. In the context of humanoid motion, such approaches suggest a practical mechanism to refine motions toward better alignment with high-level intent while respecting embodiment constraints \cite{jiang2024harmon}.
However, directly adopting language-to-humanoid motion pipelines is non-trivial for sign language. Many VLM-guided motion refinement approaches assume access to motion-centric text supervision or detailed motion descriptions \cite{jiang2024harmon}, whereas standard SL datasets typically provide spoken-language sentences paired with signing motion without explicit action descriptions (e.g., ``raise right hand to chest and rotate wrist''). This mismatch motivates VLM-guided retargeting formulations that rely on \emph{visual feedback} rather than motion-description supervision. Our work follows this emerging direction by using a VLM to critique rendered humanoid signing motion and propose targeted task-space corrections, enabling refinement under robot feasibility constraints.

% =========================
% Method (Revised for IROS)
% =========================

\section{Method}
\label{sec:method}

\subsection{Problem Definition and Overview}
\label{subsec:overview}
Given a generated signing motion represented as an SMPL-X axis--angle pose sequence
$x^{(0)} \in \mathbb{R}^{T \times 132}$, our goal is to produce a humanoid joint trajectory
$q_{1:T}$ that (i) is feasible under the robot's kinematic and geometric constraints
(e.g., joint limits and collision checking) and (ii) preserves the communicative intent
of the original signing motion.
Our pipeline consists of two stages aligned with these objectives:
(1) \textbf{collision mitigation} in the SMPL-X domain to remove hand--hand and hand--torso
interpenetrations that commonly arise in SLG outputs; and
(2) \textbf{humanoid retargeting} using an IK backbone augmented with a \textbf{VLM-guided
refinement loop} that iteratively corrects perceptually salient embodiment mismatches
via interpretable task-space control primitives.
This separation is intentional: collision mitigation addresses a systematic physical
failure mode inherited from SLG, while VLM-guided refinement targets residual
semantic/geometric discrepancies introduced by embodiment differences and IK
approximation.

\subsection{Self-Collision Mitigation via VolumetricSMPL-X}
\label{subsec:volumetric_smplx_collision}

As discussed in Sec.~I, self-penetrations are a common artifact of generators trained without explicit geometric constraints. To obtain robot-executable motion, we explicitly penalize self-intersections
using \emph{VolumetricSMPL-X} \cite{mihajlovic2025volumetricsmpl}, which augments
SMPL/SMPL-X with differentiable signed-distance-field (SDF) queries and collision-related
losses.

\textbf{Volumetric body model.}
Given a pose sequence $x \in \mathbb{R}^{T \times 132}$, we instantiate an SMPL-X model and
attach volumetric functionality using VolumetricSMPL-X~\cite{mihajlovic2025volumetricsmpl}.
This enables evaluating a differentiable self-collision objective, which penalizes surface points that penetrate the volume of other body parts:
\begin{equation}
\mathcal{L}_{\text{col}}(x)=\frac{1}{T}\sum_{t=1}^{T}\,\sum_{v \in \mathcal{V}_t} \max\big(0,\, -f\big(v;\, x_t\big)\big)^{2},
\label{eq:self_collision_loss}
\end{equation}
where $f(\cdot\,;x_t)$ denotes the differentiable signed-distance field induced by VolumetricSMPL-X for the body posed at frame $t$, and $\mathcal{V}_t$ is the set of query points sampled on the posed body surface, with each point evaluated against the SDF of the body parts it does not belong to. Negative signed distances correspond to interpenetration, so the loss is zero for collision-free configurations and grows quadratically with penetration depth, providing smooth gradients for correction. We use the loss implementation provided by VolumetricSMPL-X~\cite{mihajlovic2025volumetricsmpl}.

\textbf{Test-time collision correction by constrained optimization.}
We apply collision correction as a lightweight post-processing step on generated sequences.
Starting from the generator output $x^{(0)}$, we optimize a corrected sequence $x$ using
Adam \cite{kingma2014adam} while restricting updates to a subset of pose dimensions.
This masking prevents global drift and keeps unaffected joints unchanged:
\begin{equation}
x_{\text{eff}} \;=\; x^{(0)} + M \odot (x - x^{(0)}),
\label{eq:masked_update}
\end{equation}
where $M\in\{0,1\}^{T\times 132}$ is a binary mask selecting optimizable dimensions.

We minimize a weighted sum of (i) volumetric self-collision, (ii) a closeness prior to
preserve the original motion, and (iii) a temporal smoothness term based on acceleration:
\begin{equation}
\min_{x}\;\;
\lambda_{\text{col}} \,\mathcal{L}_{\text{col}}(x_{\text{eff}})
\;+\;
\lambda_{\text{close}} \,\lVert x_{\text{eff}} - x^{(0)} \rVert_2^2
\;+\;
\lambda_{\text{smooth}} \,\lVert \Delta^2 x_{\text{eff}} \rVert_2^2,
\label{eq:collision_objective}
\end{equation}
where $\Delta^2 x_t = x_{t+1}-2x_t+x_{t-1}$ denotes discrete acceleration.

\subsection{IK Retargeting with VLM-Guided Refinement}
\label{subsec:vlm_retargeting}

\begin{figure}[!t]
    \centering
    \begin{minipage}[t]{0.49\textwidth}
        \centering
        \includegraphics[width=\linewidth]{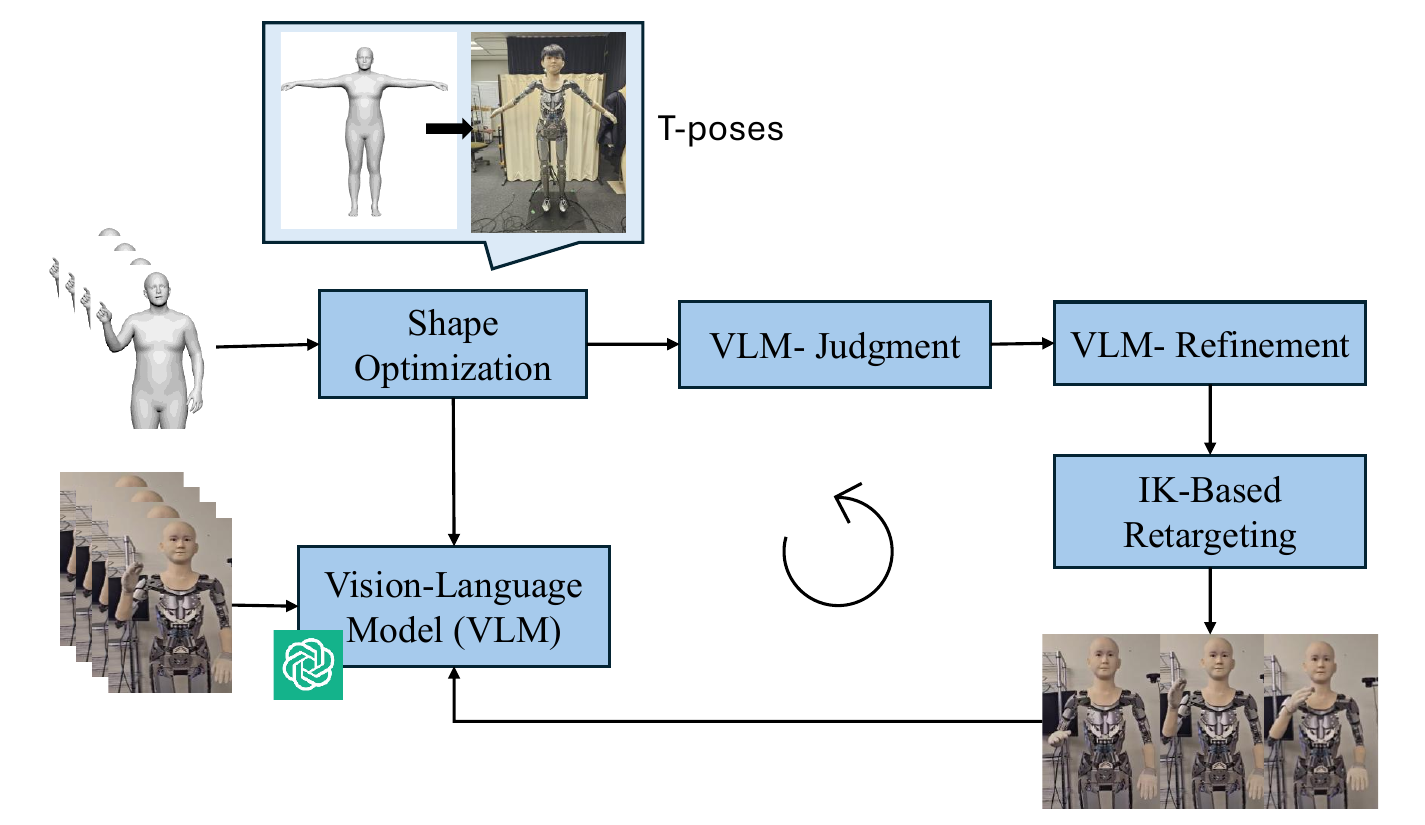}\\
        \small\textbf{(a)} Pipeline overview.
    \end{minipage}
    \hfill
    \begin{minipage}[t]{0.49\textwidth}
        \centering
        \includegraphics[width=\linewidth]{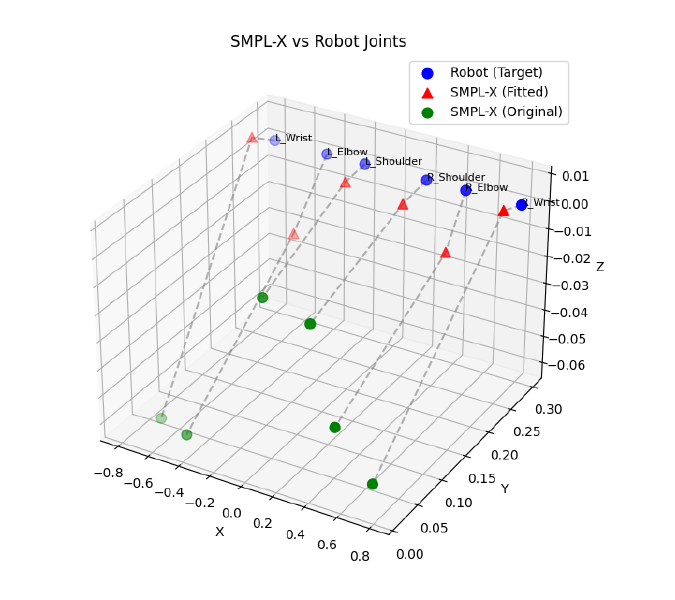}\\
        \small\textbf{(b)} Shape optimization (SMPL-X vs. robot joints).
    \end{minipage}
    \caption{Overview of the proposed pipeline VLM-guided retargeting. \textbf{(a)} A generated SMPL-X signing sequence is first corrected by an offline volumetric self-collision mitigation stage, then retargeted to the humanoid with IK. The resulting robot-rendered motion is compared against the reference SMPL-X motion by a vision–language model (VLM), which identifies embodiment-specific failures and proposes interpretable task-space correction primitives (e.g., wrist position or yaw adjustments), after which IK is re-solved. \textbf{(b)} Shape optimization used to align the SMPL-X template to the target robot embodiment before motion retargeting.}
    \label{fig:vlm_retarget_main}
\end{figure}

Figure~\ref{fig:vlm_retarget_main} summarizes our retargeting pipeline, which consists of
(1) shape optimization to align the SMPL-X body to the target robot embodiment,
(2) IK-based retargeting to obtain a feasible joint trajectory, and
(3) VLM-guided refinement in a closed loop.

\textbf{Shape optimization (SMPL-X to robot embodiment).}
A major source of retargeting error is the mismatch between the human body shape assumed by
SMPL-X and the physical dimensions of the target robot (e.g., link lengths and joint locations).
To reduce this mismatch, we perform a shape fitting step that aligns an SMPL-X template to the
robot's kinematic structure using T-pose correspondences.
Let $J^{\text{rob}} \in \mathbb{R}^{K \times 3}$ denote a set of robot joint locations in a canonical T-pose
(e.g., shoulders, elbows, wrists), and let $J^{\text{smplx}}(\beta) \in \mathbb{R}^{K \times 3}$ denote the
corresponding SMPL-X joints produced by the model as a function of SMPL-X shape parameters $\beta$
(with all pose parameters fixed to the canonical SMPL-X T-pose).
We optimize $\beta$ to match these joints by minimizing:
\begin{equation}
\min_{\beta}\;\; \mathcal{L}_{\text{shape}}(\beta)
=
\sum_{k=1}^{K} w_k \left\| J^{\text{smplx}}_k(\beta) - J^{\text{rob}}_k \right\|_2^2
\;+\; \lambda_\beta \|\beta\|_2^2,
\label{eq:shape_opt}
\end{equation}
where $w_k$ are optional joint weights (e.g., emphasizing upper-body joints relevant to signing) and
$\lambda_\beta$ regularizes the shape parameters.
We solve Eq.~\eqref{eq:shape_opt} using the Adam optimizer.
Figure~\ref{fig:vlm_retarget_main}(b) visualizes the alignment effect.

\textbf{IK-based retargeting.}
After shape alignment, we retarget the SMPL-X motion to the robot by solving inverse kinematics (IK)
over time. Given per-frame SMPL-X targets (e.g., wrist positions and orientations in the aligned body frame),
IK computes a robot joint trajectory $q_{1:T}$ that best satisfies these targets while respecting joint limits,
and we additionally perform robot-geometry collision checking (and other constraints when available, e.g.,
velocity limits). This produces a physically feasible motion, but it can still deviate from the intended signing
due to embodiment differences and IK approximation errors.

\textbf{VLM-guided refinement loop.}
To further close the semantic and geometric gap, we introduce a VLM-based refinement loop that iteratively
edits the retargeted motion. At each iteration, we render (i) a reference video from the original SMPL-X motion
(ground truth in our pipeline) and (ii) a robot-rendered video produced by IK retargeting. We provide both
videos (or corresponding frame sequences) to a vision--language model (VLM) for \textit{comparison}, and the VLM
acts as a judgment agent that assesses whether the robot motion matches the reference motion sufficiently.

\textbf{Control primitives for interpretable edits.}
Directly asking the VLM to edit robot joint angles is ineffective, because the mapping from joint configurations
to perceived motion is non-intuitive. Instead, we define a small set of \textbf{control primitives} that operate in task
space on the end-effectors (wrists), allowing the VLM to propose interpretable adjustments while the IK solver
maintains feasibility. Unless stated otherwise, offsets are defined in a fixed robot coordinate frame (e.g., the robot
base frame or a torso-aligned frame), and applied to the target wrist trajectories before re-solving IK.
We separate primitives into (i) position primitives and (ii) orientation primitives:

\begin{itemize}
    \item \textbf{Position primitives}:
    \begin{itemize}
        \item \texttt{move\_left\_up}: move the left wrist position by $+10$ cm along the robot's $+y$ direction.
        \item \texttt{move\_left\_down}, \texttt{move\_right\_up}, \texttt{move\_right\_down}, \texttt{move\_left\_in},
        \texttt{move\_left\_out}, etc.
    \end{itemize}

    \item \textbf{Orientation primitives}:
    \begin{itemize}
        \item \texttt{left\_yaw\_in}: rotate the left wrist yaw inward by $+20^\circ$.
        \item \texttt{left\_yaw\_out}, \texttt{right\_yaw\_in}, \texttt{right\_yaw\_out}, etc.
    \end{itemize}
\end{itemize}

A single refinement step is represented as a structured action list (e.g., JSON-like), which the VLM can compose
by selecting multiple primitives:
\begin{align}
\texttt{"primitives\_positions"} &: [\texttt{"move\_left\_up"}], \nonumber \\
\texttt{"primitives\_orientation"} &: [\texttt{"left\_yaw\_in"}]. \label{eq:primitive_actions}
\end{align}

\textbf{Iterative refinement procedure.}
Given the primitive actions from Eq.~\eqref{eq:primitive_actions}, we apply the corresponding task-space offsets
to the target wrist trajectories, re-run IK retargeting to obtain an updated robot motion, and render a new video.
This updated video is then fed back to the VLM judgment agent for re-evaluation. We repeat this process until the
judgment agent confirms that the retargeted motion sufficiently aligns with the reference SMPL-X motion, or a maximum
number of refinement rounds is reached. In practice, this loop provides a practical interface for VLM-driven motion
correction while maintaining robot feasibility through the IK solver.

\textbf{Deployment scope: offline pre-computation.}
The full pipeline, including collision mitigation and the VLM refinement loop, is designed as an \emph{offline} pre-processing stage rather than a real-time controller. Given the inference latency of current VLMs and the iterative render--critique--re-solve structure of the loop, we envision the pipeline being run once per sentence or per sign inventory to produce validated, robot-specific joint trajectories; at interaction time, the robot executes these pre-computed trajectories with a conventional real-time controller. VLM latency therefore does not lie on the live human--robot interaction path. Enabling online refinement, e.g., through learned approximations of the critic, is left to future work (Sec.~\ref{sec:conclusion}).

\section{Experiments}

\subsection{Experimental Setup}
We conduct experiments on the widely used SL benchmark dataset \textbf{CSL-Daily}~\cite{zhou2021improving}, a large-scale Chinese Sign Language dataset with paired spoken-language sentences and sign videos from daily conversations (18K training, 1K validation, 2K test). For all experiments, we represent signing as SMPL-X pose sequences and use the curated preprocessing released by~\cite{zuo2025signs}, which provides SMPL-X motion features with 6D rotations and 10 shape parameters. For collision mitigation, we optimize the generated SMPL-X pose parameters at test time using VolumetricSMPL-X~\cite{mihajlovic2025volumetricsmpl} and minimize Eq.~\ref{eq:collision_objective} with Adam (learning rate $1\times10^{-2}$, 500 steps). The collision-mitigation stage is applied offline to each generated sequence before retargeting (cf.\ Sec.~\ref{subsec:vlm_retargeting}). Unless stated otherwise, we optimize hand pose parameters, set $\lambda_{\text{col}}{=}50$, and use regularization weights $\lambda_{\text{close}}{=}1\times10^{-4}$ and $\lambda_{\text{smooth}}{=}1\times10^{-4}$; we report an ablation over these regularizers in Table~\ref{tab:collision_ablation}. All results are computed on the same evaluation protocol described in Sec.~\ref{subsec:quant_results}, and runtimes are reported per sequence. For VLM-based refinement, we use GPT-5.2 as the vision-language model with a fixed critique prompt to compare rendered humanoid motions against the SMPL-X reference and propose corrective task-space primitives. In all experiments, we perform two refinement iterations; empirically, we observed that two rounds were often sufficient to correct the main visible embodiment-induced deviations.

\subsection{Evaluation Metrics}
We evaluate collision mitigation using the following metrics computed for each generated pose sequence.
\textbf{Collision energy} ($\mathcal{L}_{\text{col}}$) is the VolumetricSMPL-X self-intersection loss (Eq.~\ref{eq:self_collision_loss}); we report its \textbf{Before} and \textbf{After} values aggregated across sequences (mean/median).
\textbf{Improved sequences (\%)} (\textit{Succ.}) denotes the fraction of sequences for which the final collision energy is lower than the initial one, i.e., $\mathcal{L}_{\text{col}}^{\text{after}} < \mathcal{L}_{\text{col}}^{\text{before}}$.
\textbf{Collision reduction (\%)} (\textit{Red.}) is computed per sequence as
$\frac{\mathcal{L}_{\text{col}}^{\text{before}}-\mathcal{L}_{\text{col}}^{\text{after}}}{\lvert \mathcal{L}_{\text{col}}^{\text{before}}\rvert+\epsilon}\times 100$ (with a small $\epsilon$ for numerical stability), and we report its median across sequences.
To quantify motion preservation, we report \textbf{Pose MSE}, the mean-squared difference between the corrected and original pose parameters, $\frac{1}{T}\sum_{t=1}^{T}\lVert x_t-x^{(0)}_t\rVert_2^2$.
To quantify temporal smoothness, we report \textbf{$\Delta$Accel MSE}, the change in mean squared acceleration before vs.\ after correction, where acceleration is the discrete second-order difference $\Delta^2 x_t = x_{t+1}-2x_t+x_{t-1}$; lower values indicate less added jitter.
Finally, \textbf{Time (s)} reports the wall-clock runtime per sequence for the optimization procedure (averaged over sequences).

\section{Results and Discussion}

\subsection{Quantitative Results}
\label{subsec:quant_results}

\begin{table}[!t]
\centering
\caption{Collision mitigation results. Lower is better for collision energy and distortion metrics; runtime is reported per sequence.}
\label{tab:collision_metrics_main}
\begin{tabular}{lcc}
\toprule
Metric & Before & After \\
\midrule
Collision energy (mean)   & 3.52 & 0.94 \\
Collision energy (median) & 0.82 & 0.71 \\
\midrule
Improved sequences (\%)         & \multicolumn{2}{c}{8/9 (88.9\%)} \\
Collision reduction (median)    & \multicolumn{2}{c}{21.66\%} \\
Collision reduction (mean)      & \multicolumn{2}{c}{31.93\%} \\
Pose MSE vs.\ original (median) & \multicolumn{2}{c}{$6.80\times 10^{-4}$} \\
$\Delta$Accel MSE (median)      & \multicolumn{2}{c}{+$5.71\times 10^{-3}$} \\
Runtime per sequence (mean)     & \multicolumn{2}{c}{521.7 s} \\
\bottomrule
\end{tabular}
\end{table}

Table~\ref{tab:collision_metrics_main} reports quantitative results for our collision
mitigation module. We report collision energy before/after
optimization, the fraction of sequences that improve (final $<$ initial), the median
collision reduction percentage, and distortion/smoothness proxies (pose MSE relative to
the original motion and the change in acceleration MSE). Runtime is reported per sequence.
The mean runtime of 521.7 s per sequence reflects the offline nature of this stage (Sec.~\ref{subsec:vlm_retargeting}). Our goal in this paper is to establish that collision mitigation materially improves the quality of motions before retargeting to a humanoid.

\subsection{Qualitative Analysis}
\label{subsec:qualitative}

\textbf{Self-Collision Correction}
\label{subsec:qual_collision}

\begin{figure}[!t]
    \centering
    \includegraphics[width=0.45\textwidth]{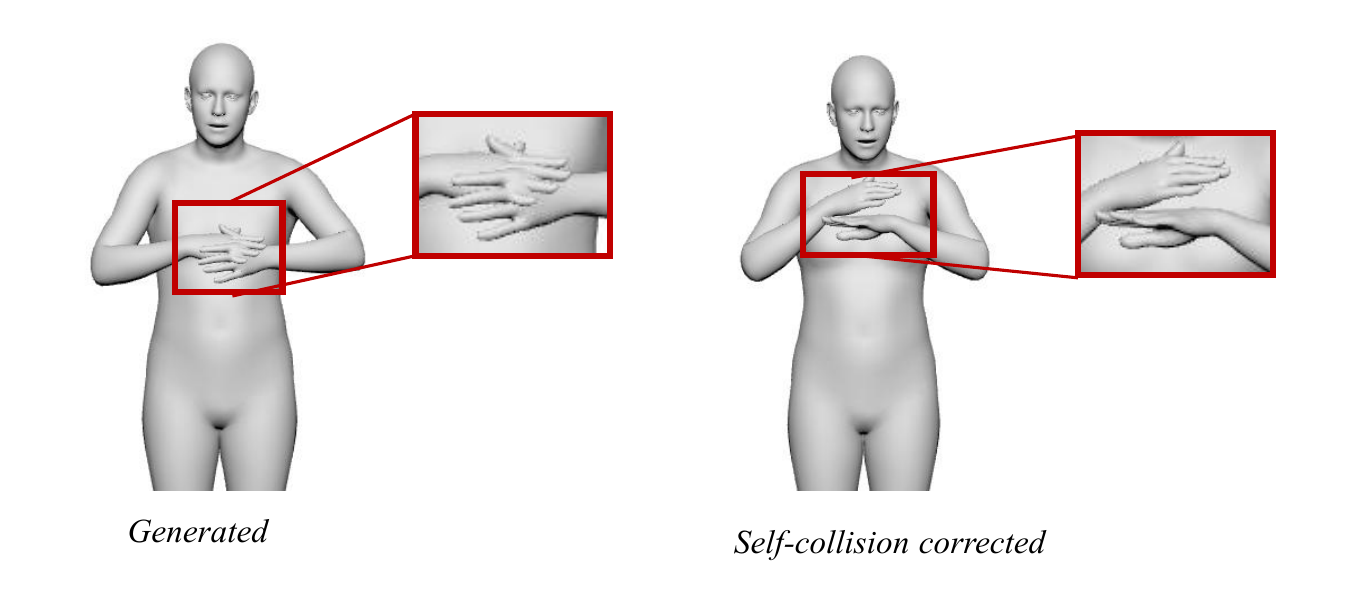}
    \caption{\textbf{Qualitative results for self-collision correction.} Left: raw generated pose exhibiting hand--hand interpenetration (zoomed inset). Right: result after VolumetricSMPL-X-based optimization, which reduces self-collision while preserving the overall signing configuration.}
    \label{fig:qual_collision}
\end{figure}

Figure~\ref{fig:qual_collision} illustrates the effect of our volumetric self-collision correction on a representative generated sequence. The left panel shows the raw generated signing pose, where the hands exhibit clear interpenetration during a bimanual interaction (highlighted in the zoomed inset). Such artifacts are common in SLG because many signs require fast, tightly coupled hand--hand or hand--body motions, while pose-space generators typically do not enforce volumetric consistency.

After applying our VolumetricSMPL-X-based optimization (right panel), the penetration is substantially reduced while the overall signing configuration is preserved. The correction primarily adjusts local relative hand placement and wrist orientation to remove self-intersection, rather than introducing large global changes in upper-body posture. This qualitative behavior is consistent with our quantitative results (Table~\ref{tab:collision_metrics_main}), which show collision reduction with limited trajectory distortion. Reducing such interpenetrations is particularly important for downstream humanoid execution, where they can lead to infeasible IK targets and unsafe motions.

\textbf{VLM-Based Retargeting}
\label{subsec:qual_vlm_retarget}

\begin{table*}[!t]
\centering
\caption{Ablation of regularization terms. ``Succ.'' denotes the percentage of sequences
with reduced collision energy (final $<$ initial), and ``Red.'' denotes the median collision reduction (\%).}
\label{tab:collision_ablation}
\begin{tabular}{lcccccc}
\toprule
Setting & $N$ & Succ.\ (\%) $\uparrow$ & Red.\ (\%) $\uparrow$ & Pose MSE $\downarrow$ & $\Delta$Accel MSE $\downarrow$ & Time (s) $\downarrow$ \\
\midrule
Collision only  & 8  & 62.5 & 18.98 & $6.05\times 10^{-4}$ & +$4.47\times 10^{-3}$ & 324.7 \\
+ Closeness  & 9  & 88.9 & 21.66 & $6.80\times 10^{-4}$ & +$5.71\times 10^{-3}$ & 521.7 \\
+ Closeness + Smoothness  & 10 & 60.0 & 24.40 & $5.08\times 10^{-4}$ & +$2.59\times 10^{-3}$ & 362.2 \\
\bottomrule
\end{tabular}
\end{table*}

\begin{figure}[!t]
    \centering
    \includegraphics[width=0.45\textwidth]{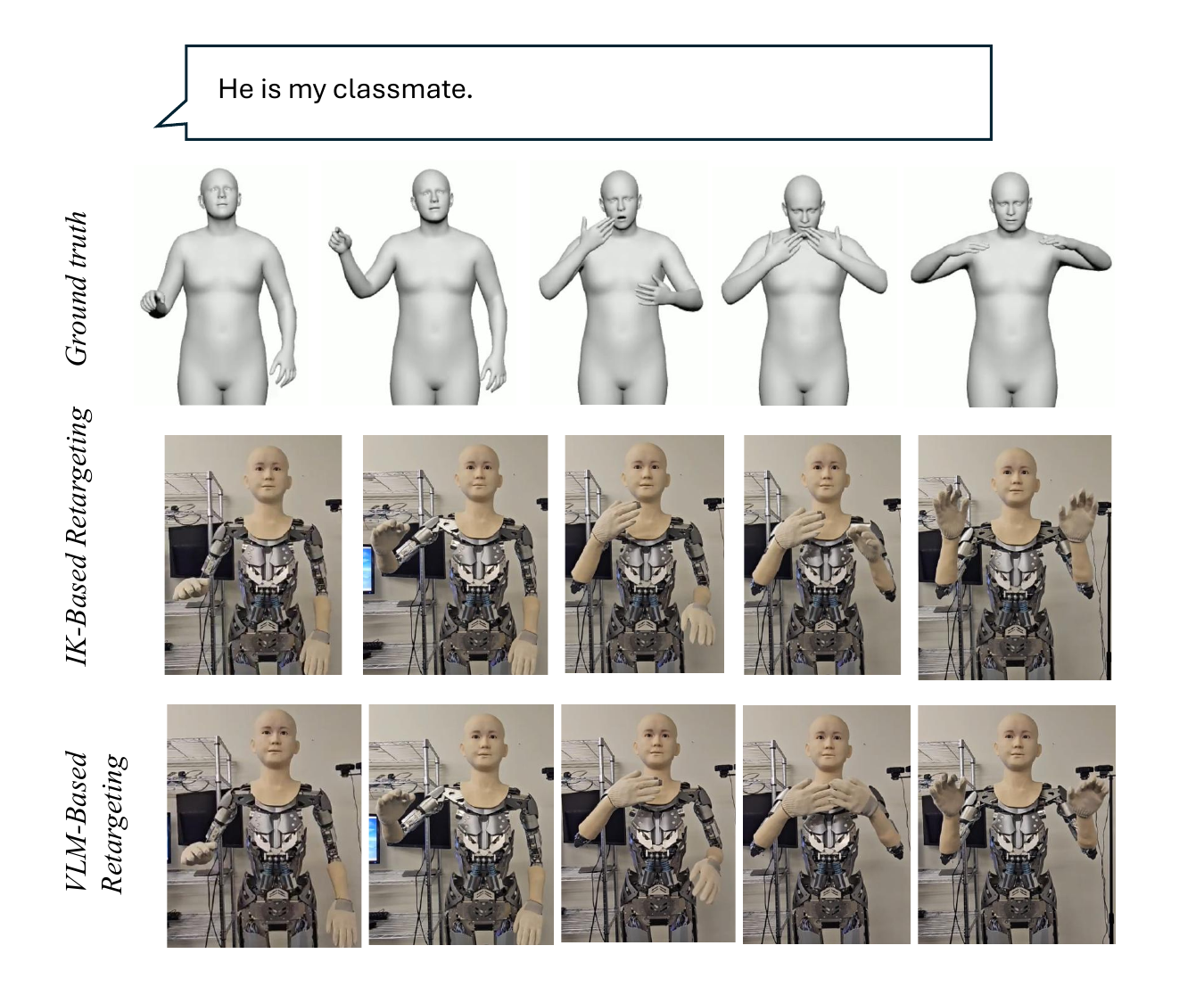}
    \caption{\textbf{Qualitative comparison of humanoid retargeting for the sentence ``He is my classmate.''}
Top: reference SMPL-X motion. Middle: IK-only retargeting, which is kinematically feasible but exhibits embodiment-induced deviation, including hand drift away from the intended signing region and mismatched wrist orientation relative to the reference motion. Bottom: VLM-refined retargeting after applying corrective task-space primitives, yielding a humanoid motion that more closely matches the reference sign realization.}
    \label{fig:vlm_retarget_qual}
\end{figure}

Figure~\ref{fig:vlm_retarget_qual} visualizes the effect of our VLM-based retargeting refinement on a representative example. We compare the reference SMPL-X motion (top row) with the robot motion produced by IK-only retargeting (middle row) and our VLM-refined retargeting (bottom row). The VLM-based refinement yields motions that are perceptually closer to the reference: hand placement relative to the torso remains within the intended signing space, wrist orientations better match the reference configuration, and the resulting motion evolution more closely follows the reference sequence.

In contrast, IK-only retargeting often produces kinematically feasible motions that exhibit systematic embodiment-induced deviations, such as end-effector offsets (hands drifting away from the target region) and orientation discrepancies that affect the perceived sign realization. By iteratively comparing the robot-rendered motion against the SMPL-X reference and applying interpretable control-primitive edits, our method reduces these mismatches without requiring motion-description supervision. We emphasize that robot experiments are shown on representative sequences; we complement these visual results with the quantitative collision-mitigation evaluation in Table~\ref{tab:collision_metrics_main}.

\subsection{Ablation Study}
\label{subsec:ablation}

Table~\ref{tab:collision_ablation} ablates the effect of regularization terms in the
collision mitigation objective. Adding a closeness prior encourages minimal deviation
from the original signing trajectory, while the smoothness term penalizes excessive
acceleration to reduce jitter. Since each run was executed on the available set of
sequences at the time of evaluation, we report $N$ for transparency.

\subsection{Sensitivity to Control-Primitive Design}
The refinement loop depends on two design choices: the primitive vocabulary and the per-primitive step sizes ($10$\,cm for position, $20^\circ$ for yaw). In our experiments, we found the vocabulary coverage to matter more than the exact magnitudes: since every proposed edit is followed by re-solving IK under joint limits and collision checking, small inaccuracies in step size are partially absorbed by the solver, whereas failure modes outside the expressible edit space (e.g., torso lean or timing errors) cannot be corrected at all. Step sizes trade convergence speed against overshoot: coarser steps correct visible offsets in fewer VLM rounds but risk oscillating around the target, while finer steps require more iterations and thus more VLM queries. Because primitives are defined in task space, the same vocabulary transfers across robot embodiments without redesign. A systematic sensitivity study over vocabulary granularity and step sizes is an important direction for future work.

\subsection{Limitations and Future Work}
The evidence supporting the VLM-guided refinement is currently qualitative and limited to representative sequences, and the collision-mitigation runtime is high (Table~\ref{tab:collision_metrics_main}). These limitations do not affect the intended offline usage of the pipeline (Sec.~\ref{subsec:vlm_retargeting}), but they bound the strength of the claims made in this paper, and we plan to address them in future work. Specifically, we will (i) quantify the benefit of VLM-guided refinement against a strong IK-only baseline using end-effector trajectory error in signing space and wrist-orientation error, (ii) scale the evaluation to a substantially larger set of sequences, and (iii) reduce the collision-mitigation runtime through learned approximations of the volumetric optimization.

\section{Conclusions}
\label{sec:conclusion}
We studied the gap between sign language generation (SLG) and humanoid execution. Although recent SLG models can generate dense 3D body representations such as SMPL-X, the resulting motions often contain self-collisions that are problematic for humanoid retargeting. To address this, we proposed a two-stage framework: an offline collision-mitigation module based on VolumetricSMPL-X that reduces self-intersection while staying close to the original trajectory, and an IK-based retargeting pipeline augmented with a vision--language model (VLM) that refines embodiment-induced errors through interpretable task-space corrections. Quantitative experiments on CSL-Daily SMPL-X sequences show reduced self-collision energy with limited distortion, while qualitative retargeting results demonstrate that VLM-guided refinement improves IK-only retargeting. Future work will investigate learned approximations or reduced-step refinement to enable online processing, a quantitative evaluation of the VLM-guided refinement against strong IK-only baselines, and validation on a physical humanoid platform.

%%%%%%%%%%%%%%%%%%%%%%%%%%%%%%%%%%%%%%%%%%%%%%%%%%%%%%%%%%%%%%%%%%%%%%%%%%%%%%%%

\section*{ACKNOWLEDGMENT}

\paragraph*{AI Assistant Disclosure}
We used large language model assistants (e.g., GPT-4 and Claude) to support this work in three ways: (i) debugging and refactoring code, (ii) improving writing clarity (grammar, wording, and organization), and (iii) serving as the vision--language model (VLM) component in our retargeting refinement loop, where the model compares rendered motions and suggests high-level control-primitive adjustments. All experimental design choices, method development, results, analyses, and scientific contributions are the authors' original work, and the final manuscript was reviewed and verified by the authors.

%%%%%%%%%%%%%%%%%%%%%%%%%%%%%%%%%%%%%%%%%%%%%%%%%%%%%%%%%%%%%%%%%%%%%%%%%%%%%%%%

\bibliography{references}
\bibliographystyle{ieeetr}

\end{document}